\documentclass[10pt,journal,cspaper,compsoc]{IEEEtran}
\usepackage{epsfig}
\usepackage{graphicx}
\usepackage{amsmath}
\usepackage{amssymb}
\usepackage{amsopn}
\usepackage{epstopdf}
\usepackage{subfigure}
\usepackage{url}
\usepackage{algorithm}
\usepackage{algorithmic}
\usepackage{multirow}
\usepackage{caption}

\ifCLASSOPTIONcompsoc
\else
\fi
\ifCLASSINFOpdf
\else
\fi

\begin{document}
\title{Tracklet Association by Online Target-Specific Metric Learning and Coherent Dynamics Estimation}

\author{Bing~Wang,~\IEEEmembership{Student Member,~IEEE,}
        Gang~Wang,~\IEEEmembership{Member,~IEEE,}
        \\Kap~Luk~Chan,~\IEEEmembership{Member,~IEEE}
        and Li~Wang,~\IEEEmembership{Member,~IEEE} 
\IEEEcompsocitemizethanks{\IEEEcompsocthanksitem B.~Wang, G.~Wang, K.~L.~Chan and L.~Wang are with the School of Electrical and Electronic Engineering, Nanyang Technological University. 50 Nanyang Avenue, Singapore, 639798. \protect\\
E-mail: \{wang0775,wanggang,eklchan,wa0002li\}@ntu.edu.sg}
\thanks{}}

\IEEEcompsoctitleabstractindextext{
\begin{abstract}
In this paper, we present a novel method based on online target-specific metric learning and coherent dynamics estimation for tracklet (track fragment) association by network flow optimization in long-term multi-person tracking. Our proposed framework aims to exploit appearance and motion cues to prevent identity switches during tracking and to recover missed detections. Furthermore, target-specific metrics (appearance cue) and motion dynamics (motion cue) are proposed to be learned and estimated online, i.e. during the tracking process. Our approach is effective even when such cues fail to identify or follow the target due to occlusions or object-to-object interactions. We also propose to learn the weights of these two tracking cues to handle the difficult situations, such as severe occlusions and object-to-object interactions effectively. Our method has been validated on several public datasets and the experimental results show that it outperforms several state-of-the-art tracking methods.
\end{abstract}

}

\maketitle
\IEEEdisplaynotcompsoctitleabstractindextext
\IEEEpeerreviewmaketitle

\section{Introduction}

\IEEEPARstart{I}{n} this paper, we address the challenges in long-term tracking of multiple persons in a complex scene captured by a single, uncalibrated camera with an aim of achieving consistent person identity tracking (i.e. no identity switches). This is a challenging problem due to many sources of uncertainty, such as clutter, serious occlusions, targets interactions, and camera motion.

Recently, significant progress has been reported in human detection \cite{Dalal,Felzenszwalb1,Felzenszwalb2,Huang,Tuzel,Wang,Wu}, and this promotes the popular tracking paradigm: detect-then-track \cite{Huang2,Li,Xing,Zhang,Berclaz,Pirsiavash,Breitenstein,Andriyenko,Butt}. The main idea is that a human detector is run on each frame to detect targets of interest, and then detection responses are linked across multi-frames to obtain target trajectories. In \cite{Zhang,Berclaz,Pirsiavash,Butt}, the authors formulate the multi-target data association as a network flow optimization problem. Zhang et al. \cite{Zhang} use a push-relabel method \cite{Goldberg} to solve the min-cost flow problem. Berclaz et al. \cite{Berclaz} and Pirsiavash et al. \cite{Pirsiavash} propose to use successive shortest path algorithms, which can achieve roughly the same tracking results with less computation cost. In a more recent paper, Butt et al. \cite{Butt}  incorporate higher-order track smoothness constraints, such as constant  velocity, for multi-target tracking. However, due to the limitation of the appearance cues used for tracking, the methods mentioned above usually cannot deal with longer term tracking to obtain a complete trajectory of a target. This is because prolonged occlusions and target-to-target interactions will result in fragmentation of a trajectory.

\begin{figure}[t]
\centering
\includegraphics[width=0.75\linewidth]{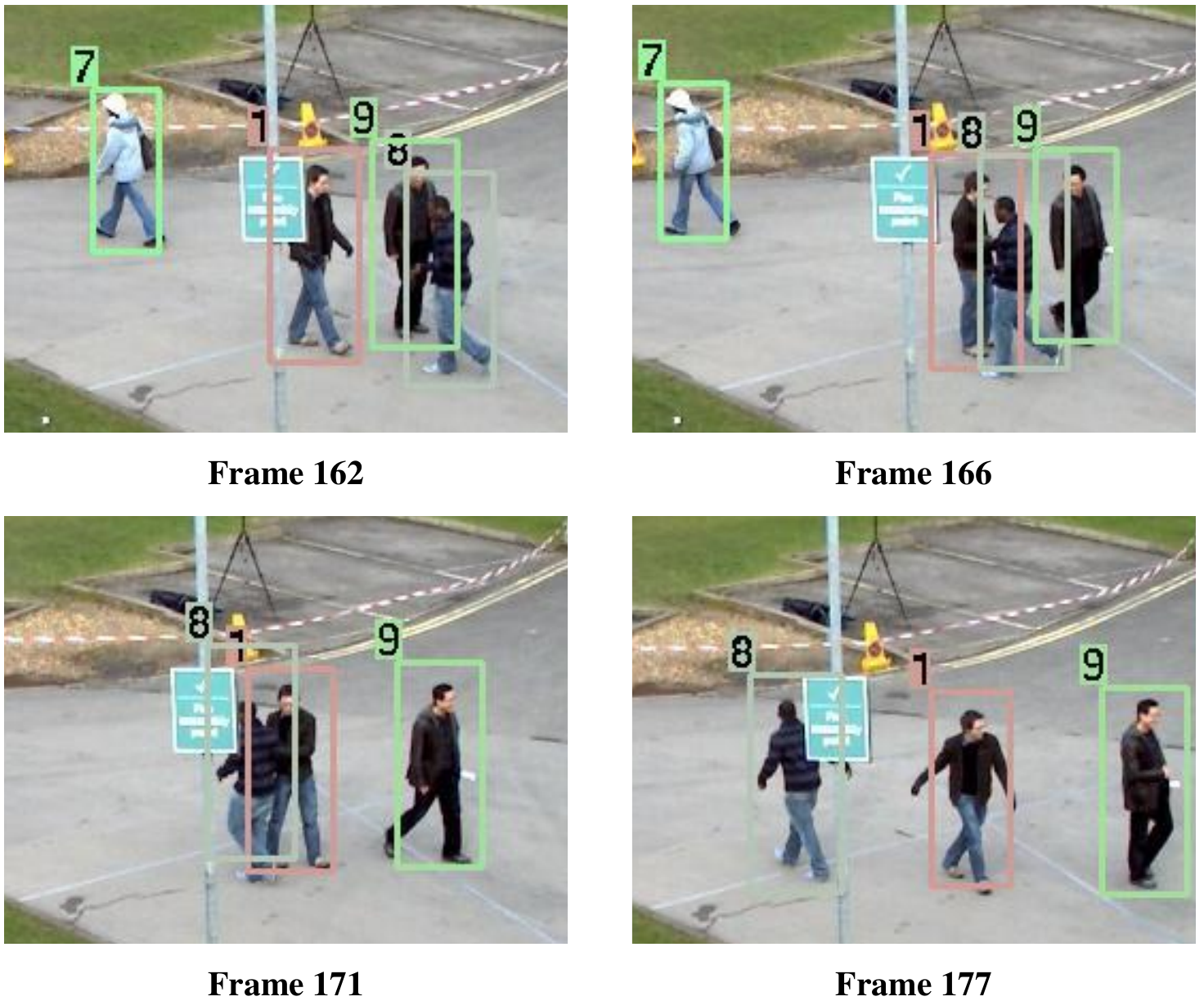}
   \caption{A difficult scenario of high appearance similarity among targets. (Frames from PETS dataset with pedestrian identities labeled by our method): Despite individuals 1 and 8 dressed in similarly colored clothes and severe occlusions and interactions between individuals 1, 8, and 9, their identities should remain unchanged.}
\label{fig:1a}
\end{figure}

\begin{figure*}[!ht]
\centering
\includegraphics[width=0.8\linewidth]{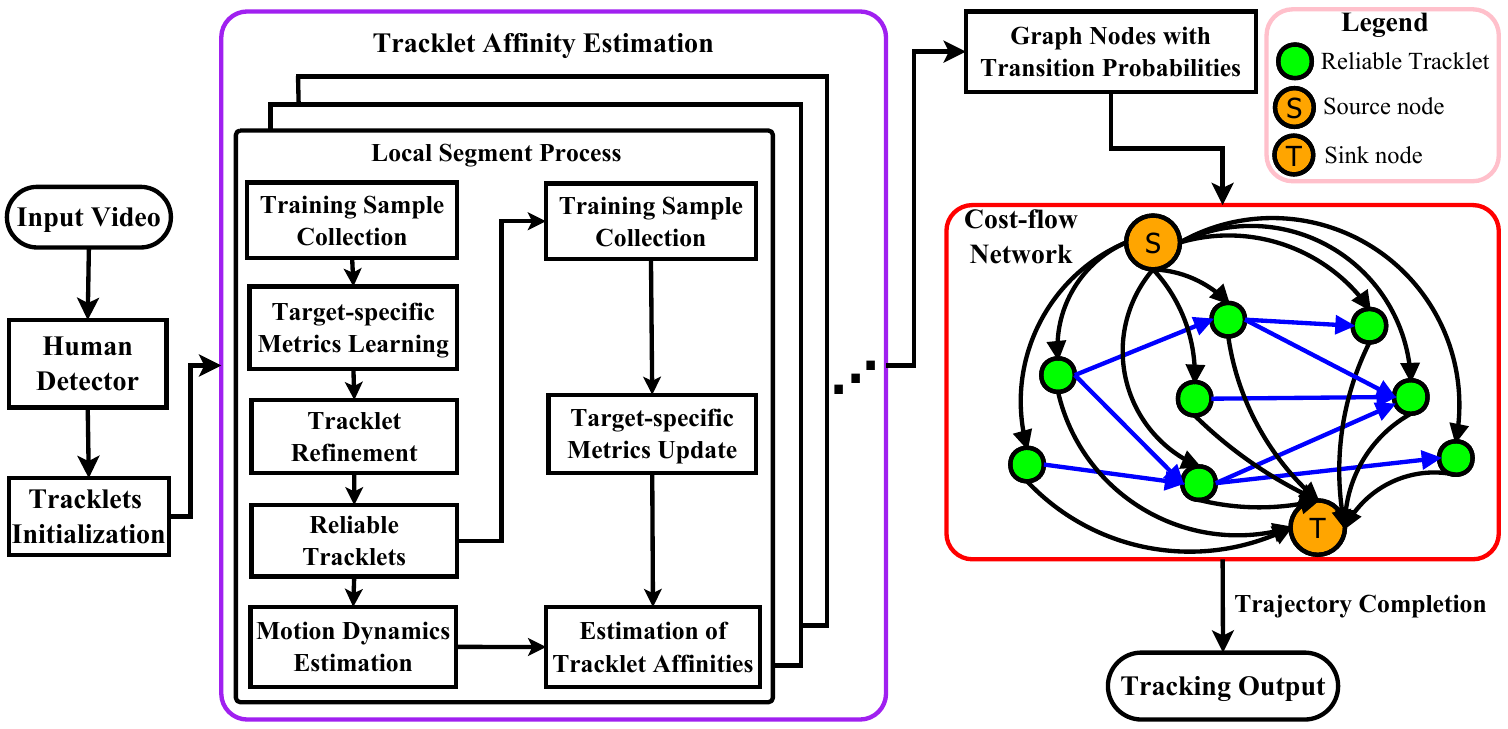}
   \caption{The proposed framework. In the cost-flow network, each node denotes a reliable tracklet, which is a tracklet with only one identity. The flow costs of  edges are defined by negative log of the affinity scores, which are obtained through the online learning of target-specific metrics and motion dynamics with the off-line learned weights on segments of short-time sequences known as local segments.}
\label{fig:1}
\end{figure*}

By using of the information from previous, current, and subsequent frames, trajectory can be recovered from the fragments and tracking errors such as missed tracks or identity switches can be corrected. In our earlier work \cite{Wang111}, we advocate a  discriminative target-specific appearance-based affinity model to reinforce the appearance cues for multi-person tracking. Unlike the PIRMPT system proposed by \cite{Kuo}, which requires off-line learned local descriptors, our target-specific metrics are online learned during tracking. In \cite{Wang111}, we utilized a motion constraint based on heuristics. In this paper, we exploit motion dynamics to further improve tracking of target's identity. Furthermore, we study the significance of the appearance and motion cues on tracking performance independently. Different from previous works \cite{Kuo,Yang3,Yang4}, which simply multiply the motion and appearance affinities to obtain the linking probabilities of two tracklets, we separately develop a learning algorithm to automatically learn the weights of the two terms from labeled training data. The learned weights can enhance the tracking cues with strong discriminative power and suppress the tracking cues with weak discriminative power. As a result, the weighted tracking cues can disambiguate the targets' respective identities better even in situations such as the one depicted in Figure \ref{fig:1a}.

A typical way of implementing the popular ``detect-then-track" paradigm is to track multiple targets frame by frame, which often encounters irrecoverable errors if a target is undetected in one or more successive frames or if two detections are erroneously linked. To overcome this weakness, global trajectory optimization over batches of frames have been proposed in recent years, using methods such as Linear Programming \cite{Storms,Jiang} and Dynamic Programming \cite{Fleuret,Pirsiavash}. These methods are often based on graphical network optimization in which the nodes are represented by detection responses. Such methods often fail to handle the problems of long-term tracking in crowded scenes well. To alleviate this, some researchers \cite{Kuo,Yang2,Shitrit01} try to use the track fragments (tracklets) as graph nodes aiming at linking tracklets into long trajectories. This kind of Tracklet Association-based Tracking (TAT) methods can increase robustness and reduce the computation complexity of the graph optimization. There are two key components of a TAT approach: (1) The tracklet affinity model that estimates the likelihood of two tracklets belonging to the same target;  (2) The global optimization framework for tracklet association that determines the links of the tracklets based on their affinity scores.

In this paper, we report our algorithm applied to tracking pedestrians in real scenes, but it can be generalized to tracking any other objects in diverse situations. The framework of this approach is shown in Figure \ref{fig:1}.

Given a video sequence, we first detect pedestrians in each frame by an existing detector, such as the Deformable Part Models (DPM) detector \cite{Felzenszwalb1}.
We utilize the strategy introduced in Section \ref{sec:3.1} to generate the initial tracklets, which are mostly reliable. But some errors, though very little, could still exist in initial tracklets. We introduce our target-specific metric learning on these initial tracklets. Then we use the online learned target-specific metrics to refine these initial tracklets for reliable tracklets. The cost-flow network is based on the reliable tracklets and its optimization yields the long-term trajectories of multiple persons. Estimating the transition costs is the key factor in the min-cost network flow optimization. We propose to learn tracklet affinity models, which include weighted discriminative appearance and motion cues, in an online manner for estimating the transition costs.

The main contributions of this paper are: (1) Online learning of target-specific metrics with strong discriminative power through a two-step target-specific metric learning and metric refinement processes. (2) Utilizing both appearance and motion dynamics in the tracklet affinity models, which are updated within each local segment for reduced computation and locally adaptive affinity models. (3) A learning algorithm to learn the weights of motion and appearance tracking cues for tracklet affinity models.

The rest of this paper is organized as follows: Section 2 describes the cost-flow network formulation. Section 3 presents the online learning of the tracklet affinity models. The learning of weights is presented in section 4. Experimental results and comparisons are shown in section 5. Section 6 concludes the paper.

\section{Cost-flow Network Formulation for Trajectory Recovery by Tracklet Association} \label{sec:2}

The cost-flow network has been shown to be effective for estimating trajectories in previous studies \cite{Zhang,Berclaz,Pirsiavash}. However, in these works, the graph nodes are defined by the detection responses. In recent works \cite{Ge,Singh,Xing,Kuo,Yang2,Yang3}, Tracklet Association-based Tracking (TAT) methods were proposed for multi-target tracking. In these methods, the tracklets were generated based on association of detection responses. In this paper, we generate the initial tracklets based the successive shortest path algorithm of \cite{Pirsiavash}, which will be described in sub-section \ref{sec:3.1}. The initial tracklets are then refined by the proposed online learned target-specific metrics for reliable tracklets. This tracklet refinement process will be presented in sub-section \ref{sec:3.3}. We can construct a smaller graph based on such tracklets which are of a higher order of abstraction than those based on detection responses. The problems in long-term multi-person tracking can be solved by directly linking tracklets instead of detection responses.

An objective function, which takes a similar form as detection association  in \cite{Zhang}, is defined for tracklet association. Let $X=\{F_i\}$ be the collection of all the tracklets. A single trajectory hypothesis is defined as an ordered list of $N$ tracklets: $T_k=\{F_{k_1},F_{k_2},...,F_{k_l}\}$, where $F_{k_i}\in X$, and $i=1,...,l; 1\leq l<N$. A tracklet association hypothesis $\mathcal{T}$ is defined as a set of single trajectory hypotheses: $\mathcal{T}=\{T_k\}$.

The objective of tracklet association is to maximize the posteriori probability of $\mathcal{T}$ given $X$:
\vspace*{-0.5\baselineskip}
\begin{align}
\mathcal{T^{*}} &= \arg\max_\mathcal{T} P(\mathcal{T}|X) \notag\\
& = \arg\max_\mathcal{T} P(X|\mathcal{T})P(\mathcal{T})  \notag\\
& = \arg\max_\mathcal{T} \prod_i P(F_i|\mathcal{T})P(\mathcal{T}) \label{eq:1}
\end{align}
assuming that the likelihood probabilities of $F_i$ are conditionally independent.

We take the assumption that the motion of each tracklet is independent and one tracklet can only belong to one trajectory. Then the above equation can be further decomposed into:
\vspace*{-0.5\baselineskip}
\begin{align}
\mathcal{T^{*}}= & \arg\max_\mathcal{T} \prod_i P(F_i|\mathcal{T})\prod_{T_k\in \mathcal{T}} P(T_k) \label{eq:2} \\
& s.t.\quad T_k\bigcap T_l=\phi, \forall k\neq l \label{eq:3}
\end{align}
where $\phi$ is the empty set.

The second term in Equ. \eqref{eq:2} is defined as follows:
\vspace*{-0.5\baselineskip}
\begin{align}
P(T_k)& = P(\{F_{k_1},F_{k_2},...,F_{k_l}\}) \notag \\
& = P_s(F_{k_1})\left(\prod\limits_{n=1}^{l-1}P(F_{n+1}|F_n)\right)P_t(F_{k_l}) \label{eq:5}
\end{align}

$P(F_i|\mathcal{T})$ is the likelihood function of tracklet $F_i$. It is assumed that the false alarm rate is very low from the reliable tracklets, so $P(F_i|\mathcal{T})\approx 1$. Then Equ. \eqref{eq:2} can be further simplified as follows:
\vspace*{-0.5\baselineskip}
\begin{align}
\mathcal{T^{*}}& = \arg\max_\mathcal{T} \prod_i P(F_i|\mathcal{T})\prod_{T_k\in \mathcal{T}} P(T_k) \notag\\
&= \arg\max_\mathcal{T} \prod_{T_k\in \mathcal{T}} P(T_k) \label{eq:6}
\end{align}

$P(T_k)$ is modeled as a Markov chain, which includes a starting probability $P_s(F_{k_1})$, a termination probability $P_t(F_{k_l})$, and transition probability $P(F_{n+1}|F_n)$ between temporarily adjacent tracklets. Finding the optimal association hypothesis $\mathcal{T^{*}}$ is equivalent to minimizing the cost of flow from source $s$ to sink $t$ in a network flow graph. A network graph can be constructed as follows:

Given an observation set $X$: for every tracklet $F_i\in X$, we create a node $v_i$, an edge from source $s$ to a node,$(s,v_i)$, with cost $c(s,v_i)=c_i^{s}$ and flow $f(s,v_i)=f_i^{s}$, and an edge from a node to sink $t$, $(v_i,t)$ with cost $c(v_i,t)=c_i^{t}$ and flow $f(v_i,t)=f_i^{t}$. For every transition $P(F_j|F_i)\neq 0$, create an edge $(v_i,v_j)$ , $i\neq j$, with cost $c(v_i,v_j)=c_{ij}$ and flow $f(v_i,v_j)=f_{ij}$.  We take the logarithm of the objective function to simplify the expression while preserving the maximum a posteriori probability (MAP) solution.
Then, Equ. \eqref{eq:6} can be re-written as follows:
\vspace*{-0.5\baselineskip}
\begin{align}
\mathrm{T}=& \arg\min_\mathcal{T} \left(\sum_i c_i^{s}f_i^{s} + \sum_{ij} c_{ij}f_{ij} + \sum_i c_i^{t}f_i^{t}\right) \label{eq:7} \\
& s.t. \quad f_{ij},f_i^{s},f_i^{t}\in \{0,1\} , \notag\\
& and \quad f_i^{s} + \sum_j f_{ji}=f_i^{t} + \sum_j f_{ij} \label{eq:8}
\end{align}
subject to Equ. \eqref{eq:7}, where
\vspace*{-0.5\baselineskip}
\begin{align}
& c_i^{s}=-\log P_s(F_i), \quad c_i^{t}=-\log P_t(F_i), \notag\\
& c_{ij}=-\log P(F_j|F_i). \notag
\end{align}

Equ. \eqref{eq:8} ensures that the tracklet association hypothesis $\mathcal{T}$ is non-overlapping. The above formulation can be mapped into a cost-flow network with a source $s$ and a sink $t$. Estimating the transition costs $c_{ij}$ is very critical in solving this min-cost network flow problem. Previous network flow approaches \cite{Zhang,Berclaz,Pirsiavash,Butt} only utilize motion cues across consecutive frames and simple appearance features such as color histograms to calculate $c_{ij}$. Nevertheless, these cues are not very reliable when prolonged occlusions and interactions between targets occur. In this paper, we propose to learn the segment-wise tracklet affinity models, consisting of weighted tracking cues, online for estimating $c_{ij}$.

\section{Online Learning of Tracklet Affinity Models}

In this section, we introduce the online learning of tracklet affinity models, consisting of online target-specific metric learning and online motion dynamics estimation. The affinity scores of adjacent tracklets, which are used as the transition probabilities between two corresponding nodes in the cost-flow network, can be obtained through tracklet affinity measurements. We perform local transition probabilities estimation within a local segment of $S$ frames ($S=50$ in our implementation).

In order to obtain effective appearance cues for reliable transition probability estimation, we propose a novel target-specific appearance-based model. The appearance-based model learning problem is formulated as a metric learning problem, which can enhance the features with strong discriminative power and suppress the features with weak discriminative power. Here, we learn target-specific metrics so that target-specific properties can be efficiently explored for more discriminative models. In contrast to the previous work of \cite{Kuo} in which local descriptors are learned offline, our learning is online throughout and our target-specific metrics are adaptive to local segments. Moreover, to create a more discriminative tracklet affinity model, we also explore the motion dynamics cue and embed it into the proposed tracklet affinity model. The motion dynamics are online estimated without any assumed priors. As a result, the learned tracklet affinity models can better represent the appearance and motion cues adaptively and provide reliable transition probability estimation.

\subsection{Initial Tracklet Generation} \label{sec:3.1}

Before introducing the proposed tracklet affinity models, we first present the initial tracklet generation process. Similar to the cost-flow network formulation described in previous section, the problem of initial tracklet generation is also formulated as a network flow optimization problem. Different from the cost-flow network formulation presented in Section \ref{sec:2}, the graph nodes are defined by the detection responses and their costs, referred as local observation costs here, (the negative logarithm of the corresponding detection scores) are added in the formulation.
The mathematical formulation of the minimization problem can be expressed as follows:
\vspace*{-0.5\baselineskip}
\begin{align}
\mathrm{F}=& \arg\min \left(\sum_i \hat{c}_i^{s}\hat{f}_i^{s} + \sum_{ij} \hat{c}_{ij} \hat{f}_{ij} + \sum_i \hat{c}_i \hat{f}_i + \sum_i \hat{c}_i^{t}\hat{f}_i^{t}\right) \label{eq:7i} \\
& s.t. \quad \hat{f}_{ij},\hat{f}_i,\hat{f}_i^{s},\hat{f}_i^{t}\in \{0,1\} , \notag\\
& and \quad \hat{f}_i^{s} + \sum_j \hat{f}_{ji}=\hat{f}_i=\hat{f}_i^{t} + \sum_j \hat{f}_{ij} \label{eq:8i}
\end{align}
where $\hat{c}_i^{s}$, $\hat{c}_i^{t}$, $\hat{c}_{ij}$, $\hat{c}_i$ denote the starting, termination, transition and local observation costs, respectively. $\hat{f}_i^{s}$, $\hat{f}_i^{t}$, $\hat{f}_{ij}$, $\hat{f}_i$ denote the corresponding flows.

To obtain the initial tracklets (relatively short track fragments), the transition cost $\hat{c}_{ij}$ is set to 0 for each candidate detection pair first. Only the detection responses over consecutive frames are considered to be linkable. The transitions of the cost-flow network across non-consecutive frames are not permitted. Moreover, a constraint is imposed on initial tracklet generation. That is, a track fragment should start from a detection response and terminate at a detection response with detection scores higher than a pre-defined threshold, thus only consecutive detection responses with detection scores above the threshold are used to form the initial tracklets.
This constraint ensures that the generated initial tracklets are relatively short and mostly reliable track fragments. The dynamic programming algorithm in \cite{Pirsiavash}, which approximates the successive shortest path solution, is employed to optimize the cost-flow network. This initial tracklet generation strategy can be viewed as a simplified version of the method in \cite{Pirsiavash}. The average speed of this process is above 500 fps.

\subsection{Online Target-Specific Metric Learning} \label{sec:3.2}

We aim to online learn discriminative target-specific metrics while keeping the computational complexity low. For each tracklet $F_i$, we learn a distance metric function.

The learning involves feature representation, online training sample collection and online training. To create a strong appearance-based model, we start from a rich set of basic features, which includes color, shape and texture, to describe a pedestrian's appearance.

Given a training dataset $Z=\{z_i^{t}\}_{i=1}^{N_z}$, where $z_i^{t} \in \mathbb{R}^{N_d}$ is a feature vector representing the appearance of the image area under the bounding box where there is a strong detection response in tracklet $F_i$ at frame $t$, $N_z$ is the total number of training samples and $N_d$ is the total number of feature dimensions. The training dataset $Z$ is obtained from reliable tracklets through online training sample collection, which will be introduced in latter part of this subsection. We define a positive difference vector $x_i^{p}$ computed between a positive sample pair (a pair of detection responses belonging to the same person) and a negative difference vector $x_i^{n}$ computed from a negative sample pair (a pair of detection responses belonging to different persons). Here, it is assumed that the first $M$ frames of each initial tracklet are reliable and the detection responses are from the same person. Training samples are therefore collected from these frames. The value of $M$ is empirically determined. It is found that a value $M$ between 6 and 10 frames works well for all sequences. Based on the constraint introduced in previous subsection, most of the generated initial tracklets are relatively short and mostly reliable. However, some unreliable tracklets may still exist due to occlusions of the targets. This kind of target interactions often occur in the middle of initial tracklets. Based on this observation, we assume that the first $M$ frames of each initial tracklet are mostly reliable.

The difference vectors $x_i^{p}$ and $x_i^{n}$ are defined as:
\vspace*{-0.5\baselineskip}
\begin{align}
& x_i^{p}=d(z_i,z_i')=|z_i-z_i'|  \notag \\
& x_i^{n}=d(z_i,z_j')=|z_i-z_j'|, \quad i\neq j \label{eq:9}
\end{align}
where $d$ is an absolute difference function, $z_i$ and $z_i'$ are the feature vectors of two samples from the same tracklet $F_i$, $z_j'$ is the feature vector of a sample from a different tracklet $F_j$.

Given the difference vectors $x_i^{p}$ and $x_i^{n}$, a distance function $D_i$ for tracklet $F_i$ can be learned based on relative distance comparison so that $D_i(x_i^{p})<D_i(x_i^{n})$. This distance function $D_i$ is parameterized as a Mahalanobis distance function:
\vspace*{-0.5\baselineskip}
\begin{align}
D_i(x)=x^{T}M_ix, \quad M_i\succeq 0 \label{eq:a12}
\end{align}
where $M_i$ is a positive semidefinite matrix.

We adopt the logistic function as in \cite{Zheng2} to learn $D_i$ to force $D_i(x_i^{p})$ to be small, and $D_i(x_i^{n})$ to be big:
\vspace*{-0.5\baselineskip}
\begin{align}
\min_{D_i} r(D_i)=-\log \left(\left( 1+\exp \left(D_i(x_i^{p})-D_i(x_i^{n})\right)\right)^{-1} \right) \label{eq:11}
\end{align}

Furthermore, the term $M_i$ in the distance function $D_i$ can be decomposed by eigendecomposition:
\vspace*{-0.5\baselineskip}
\begin{align}
M_i=A_i\Lambda_i A_i^{T}=W_iW_i^{T}, \quad W_i=A_i\Lambda_i^{\frac{1}{2}} \label{eq:12}
\end{align}
where $A_i$ is the orthonormal eigenvector matrix of $M_i$ and the diagonal of $\Lambda_i$ are the corresponding eigenvalues.

Therefore, learning a distance function $D_i$ is equivalent to learning the matrix $W_i$ as follows:
\vspace*{-0.5\baselineskip}
\begin{align}
& \min_{W_i} r(W_i), s.t. \quad w_i^{T}w_j=0, \forall i\neq j, w_i,w_j\in W_i \notag \\
& r(W_i)=\log (1+\exp \{\|W_i^{T}x_i^{p}\|^{2}-\|W_i^{T}x_i^{n}\|^{2}\}) \label{eq:13}
\end{align}

Online training sample collection is another important issue in online learning. The $q$ strongest ($q=4$ in this work) detection responses in each tracklet are used as training samples. For  $x_i^{p}$, we collect positive sample pairs from the same tracklet. However, for  $x_i^{n}$, we  collect negative sample pairs from different persons. To determine the relevance of sample pairs, two constraints are employed: spatio-temporal and exit constraints. The first constraint is based on the fact that one person cannot appear at two or more different locations at the same time. The second constraint is based on the observation that the person who has already exited the view cannot be the person who is still within the view.
We online collect negative samples, which satisfy the above two constraints, from $F_i$ and $F_j$ respectively to form negative sample pairs.

Learning $W_i$ using the optimization criterion \eqref{eq:13} is a nonconvex optimization problem. In this work, we utilize the optimization algorithm in \cite{Zheng2} to learn $W_i$ for each tracklet $F_i$. Finally, we obtain the target-specific transform matrices for all the tracklets:
\vspace*{-0.5\baselineskip}
\begin{align}
W=\{W_i\}, \quad i=1,...,N \label{eq:14}
\end{align}

\subsection{Tracklet Refinement} \label{sec:3.3}

To solve the objective function in Equ. \eqref{eq:7}, we need to identify reliable tracklets for the nodes in the network graph. The strategy of initial tracklet generation, which is described in previous sub-section, uses spatio-temporal information such as distance between corresponding observations in adjacent frames to link the detections into tracklets. Without effective use of appearance cues, the initial tracklets may be not consistent in appearance and hence unreliable when there are many interactions or occlusions between targets. A typical error is that there are some detection responses belonging to different persons in one tracklet. Hence, tracklet refinement is needed to separate tracklets into multiple short but reliable ones.

The online learned target-specific metrics are employed to refine the initial tracklets. To construct the probe set, the detection with the strongest detection response, $g_i$, is selected from the first $M$ frames of an initial tracklet, $F_i$, which are assumed to be reliable. It is defined as $G=\{g_i\}$, $i=1,...,N_s$, where $N_s$ is the number of tracklets in a local segment. Each tracklet $F_i$ has only one selected $g_i$  in $G$.

We learn the target-specific transform matrix $W_i$ for each initial tracklet after collecting training samples as described in previous sub-section. The identity test is carried out within a local segment frame by frame to obtain the relative distance between the detection response at frame $t$ of $F_i$ and the corresponding $g_i$ in the probe set:
\vspace*{-0.5\baselineskip}
\begin{align}
& x_i^{t}=d(z_i^{t}-g_i)=|z_i^{t}-g_i|; \quad i=1,...,N_s  \notag \\
& d_i^{t}=\|W_i^{T}x_i^{t}\|^{2} \label{eq:17}
\end{align}
where $z_i^{t}$ is an instance from tracklet $F_i$ at frame $t$, $g_i$ is the corresponding detection response of $F_i$ in $G$, and $d_i^{t}$ is the relative distance between $z_i^{t}$ and $g_i$.

To be a reliable tracklet, the relative distance between the current detection response $z_i^{t}$ and the probe $g_i$ should be small; otherwise, it is an unreliable tracklet.  A distance threshold $\omega$ is used to identify reliable tracklets. In a tracklet $F_i$, if $K$ ($K=5$ in our implementation) consecutive detection responses having relative distance values (from $g_i$) above $\omega$, we split $F_i$ into two parts from  the first consecutive detection response.
In virtue of the strategy of initial tracklet generation introduced in Section \ref{sec:3.1}, the generated initial tracklets are mostly reliable, but some errors, though very little, could still exist in the initial tracklets. This tracklet refinement process is usually repeated no more than twice to obtain reasonably reliable tracklets.

After obtaining the reliable tracklets, the target-specific metrics learned from the initial tracklets are updated. Different from the training sample collection of initial tracklets that the samples are collected from the first $M$ frames of the tracklets, the training samples of the reliable tracklets are collected from the full-length tracklets. The strategy of online training sample collection is the same as the one introduced in Section \ref{sec:3.2}. As shown in Figure \ref{fig:1}, a two-step target-specific metric learning/update is used in the proposed framework. The first step is used for tracklet refinement, which is usually repeated no more than two times. The second step is used for the appearance-based tracklet affinity estimation, which executes only once.

\subsection{Online Tracklet Dynamics Estimation}

To disambiguate targets with similar appearance, we propose to exploit motion dynamics together with appearance cues as described above to keep track of target's identity. The main idea of tracklet dynamics estimation is to model the evolution of target motions as a sequence of piecewise linear regressors whose orders can be estimated from available data.

The dynamics of a tracklet can be constructed as an ordered sequence of dynamic measurements $\{y_{q}\}$, $s\leq q \leq e$, where $s$ and $e$ are the starting and ending frames, respectively. Similar to \cite{Camps,Ding}, we collect the position information of all the detection responses within one tracklet in a vector $y$ and assume that its value at current time $q$ is related to its past values $y_{q-i}$ by an $m^{th}$ order autoregressive model of the form:
\vspace*{-0.5\baselineskip}
\begin{align}
y_{q} &=a_1 y_{q-1} + a_2 y_{q-2} + ... + a_m y_{q-m} \notag \\
&=\sum_{i=1}^m a_i y_{q-i}, \quad m\leq N_f, q\geq s+m \label{eq:17a}
\end{align}
where $\textbf{a}={[a_1 \ a_2 \ ... \ a_m ]}^T$ is the regressor vector, $N_f$ is the total number of frames of the tracklet, $m$ is the number of frames of the dynamic measurement $y_{q}$ and $s$ is the starting frame of the tracklet.

The order of the autoregressive model $m$ measures the complexity of the underlying tracklet dynamics. The goal of tracklet dynamics estimation is, given dynamic measurements $\{y_{q}\}$, to estimate the minimum $m$ such that the model \eqref{eq:17a} retains. Specifically, a well known result from the realization theory \cite{Ho,Moonen} is that, under mild conditions, given an ordered sequence of measurements $\{y_{q}\}$ generated by Equ. \eqref{eq:17a}, the order $m$ of the autoregressive model equals to the rank of the corresponding Hankel matrix, \emph{i.e.}, $m=rank(H_{F_i})$ where $H_{F_i}$ is the Hankel matrix with $n\geq m$ columns:
\vspace*{-0.5\baselineskip}
\begin{align}
H_{F_i}\doteq
\left[
 \begin{array}{cccc}
             y_{s},  & y_{s+1} & \ldots & y_{s+n-1} \\
             y_{s+1} & y_{s+2} & \ldots & y_{s+n}   \\
             \vdots  & \vdots  & \vdots & \vdots    \\
             y_{t-n+1} & y_{t-n} &\ldots & y_{t}
             \end{array}
\right] \label{eq:17b}
\end{align}
where $n$ is defined based on the length of the tracklet:
\vspace*{-0.5\baselineskip}
\begin{align}
& n=l_i - \lceil l_i/3 \rceil + 1; \label{eq:17b2} \\
& l_i=t-s+1 \notag
\end{align}
where $l_i$ is the length of tracklet $F_i$.

The motion dynamics similarity $P_m(F_i,F_j)$ between two tracklets $F_i$ and $F_j$, which takes a similar form as in \cite{Ding}, is defined as follows:
\vspace*{-0.5\baselineskip}
\begin{align}
P_m(F_i,F_j) = \left\{
 \begin{array}{ll}
             -\infty, \quad \text{if temporal conflict exists}; \\
             \frac{rank(H_{F_i})+rank(H_{F_j})}{rank(H_{F_{ij}})} - 1
             \end{array}
\right. \label{eq:17c}
\end{align}
where $F_{ij}=[F_i \ {\alpha_i}^j \ F_j]$ is the joint tracklet with the gap ${\alpha_i}^j$ between $F_i$ and $F_j$ interpolated. The joint Hankel matrix $H_{F_{ij}}$ is formed by combining the dynamic measurements of $F_i$, $F_j$ and the interpolated data.

Here, we take the assumption that the targets do not significantly change their dynamics between tracklets. The intuition of the above motion dynamics similarity is that if two tracklets are from the same trajectory then they can be approximated by one relatively low order regressor. Otherwise, if two tracklets are from different trajectories, the joined trajectory needs a higher order regressor than the regressors of each single tracklet. Hence, if $rank(H_{F_i})=r(F_i)$ and $rank(H_{F_j})=r(F_j)$, then $rank(H_{F_{ij}})=r(F_{ij})\leq (r(F_i)+r(F_j))$. Consequently, if $F_i$ and $F_j$ are of the same trajectory, then $r(F_i)=r(F_j)=r(F_{ij})$ and $P_m(F_i,F_j)=1$, but if not, $P_m(F_i,F_j)\approx 0$.

Tracklet dynamics are online estimated, without any prior knowledge, based on the reliable tracklets. The IHTLS (Iterative Hankel Total Least Squares) method in \cite{Dicle} is employed to estimate the rank of the Hankel matrices for tracklet dynamics estimation. The computational complexity of rank estimation is $O((l_i-m)m^3)$, where $l_i$ is the length of a tracklet $F_i$ and $m \ll l_i$ is the rank of the matrix.

\subsection{Tracklet Affinity Measurement}

In this subsection, we present the measurement of the affinity between $F_i$ and $F_j$, or equivalently, the transition probability, $P_{ij}$, in the network graph between node $i$ and node $j$. The tracklet affinity score, $\mathcal{S}_{ij}$, which is equivalent to $P_{ij}$, is defined as follows:
\vspace*{-0.5\baselineskip}
\begin{align}
\mathcal{S}_{ij}=P_m(F_i,F_j)P_a(F_i,F_j)\mathcal{C}_{ij}  \label{eq:17d}
\end{align}
where $P_m(F_i,F_j)$ is the motion-based affinity model, which is defined by Equ. \eqref{eq:17c}, $P_a(F_i,F_j)$ is the appearance-based affinity model and  $\mathcal{C}_{ij}$ is a limiting function.

To obtain the appearance-based affinity model $P_a(F_i,F_j)$, we first compute the relative distances $d_{ij}^{t}$ between each detection response in $F_i$ and the probe $g_j$, and $ d_{ji}^{t'}$ between each detection response in $F_j$ and the probe $g_i$,
\vspace*{-0.5\baselineskip}
\begin{align}
& x_{ij}^{t}=|z_i^{t}-g_j|, \ x_{ji}^{t'}=|z_j^{t'}-g_i|; \quad i,j=1,...,N_s \notag \\
& d_{ij}^{t}=\|W_i^{T}x_{ij}^{t}\|^{2}, \ d_{ji}^{t'}=\|W_j^{T}x_{ji}^{t'}\|^{2} \label{eq:18a}
\end{align}
where $z_i^{t}$ denotes the feature vector of a detection response in tracklet $F_i$ at frame $t$, $z_j^{t'}$ denotes the feature vector of a detection response in tracklet $F_j$ at frame $t'$, and $g_i,g_j\in G$.

Subsequently, we calculate the mean values of the relative distances and use them to define the appearance-based affinity model $P_a(F_i,F_j)$:
\vspace*{-0.5\baselineskip}
\begin{align}
&d_{ij}=(\sum_t d_{ij}^{t})/m, \ d_{ji}=(\sum_{t'} d_{ji}^{t'})/n \label{eq:18b}\\
& P_a(F_i,F_j)=(d_{ij}d_{ji})^{-1}\gamma  \label{eq:18}
\end{align}
where $\gamma$ is a normalization term and $m$, $n$ are the number of frames of $F_i$ and $F_j$ respectively.

We do not have to apply Equ. \eqref{eq:17d} to every pair, since there are a lot of obviously non-related tracklet pairs which do not belong to the same trajectory. Because a limiting function $\mathcal{C}_{ij}$ is included in Equ. \eqref{eq:17d}, we actually apply it to every tracklet pair. This limiting function $\mathcal{C}_{ij}$ is proposed based on spatio-temporal, and exit constraints:
\vspace*{-0.5\baselineskip}
\begin{align}
\mathcal{C}_{ij}=C_t(F_i,F_j)C_e(F_i,F_j) \label{eq:19}
\end{align}

The spatio-temporal constraint is defined as follows:
\vspace*{-0.5\baselineskip}
\begin{align}
C_t(F_i,F_j) = \left\{
 \begin{array}{ll}
             1, & \text{if} \ F_i\cap F_j = \phi \\
             0, & \text{otherwise}
             \end{array}
\right. \label{eq:15}
\end{align}
where $\cap$ is an intersection operator that is used to find the overlap between two tacklets and $\phi$ is the empty set.

The exit constraint is defined based on the observation that the person who has already exited the scene cannot be the person who is still within the scene:
\vspace*{-0.5\baselineskip}
\begin{align}
C_e(F_i,F_j) = \left\{
 \begin{array}{ll}
             1, & \text{if} \ t_i^{s}>t_j^{e} \ \& \ p_j^{t_j^{e}} \notin E \\
             0, & \text{otherwise}
             \end{array}
\right. \label{eq:16}
\end{align}
where $t_i^{s}$ is the starting frame of tracklet $F_i$, $t_j^{e}$ is the ending frame of tracklet $F_j$, $p_j^{t_j^{e}}$ is the position of the detection response of tracklet $F_j$ at time $t_j^{e}$ and $E$ is the exit area which is near image borders. For static cameras, we adopt the incremental learning algorithm for exit map as in \cite{Yang3} to obtain $E$.

$C_t(F_i,F_j)$ and $C_e(F_i,F_j)$ associate $F_i$ and $F_j$ if they have no overlap and $F_i$ does not exit the screen when $F_j$ appears.

The transition costs of the adjacent nodes in the cost-flow network is obtained by taking negative logarithm of the affinity scores between corresponding tracklets:
\vspace*{-0.5\baselineskip}
\begin{align}
c_{ij}=-\log \mathcal{S}_{ij} \label{eq:21}
\end{align}

Finally, we can estimate the optimal tracklet association hypothesis $\mathcal{T^*}$ in Equ. \eqref{eq:7} based on $c_{ij}$. Although the transition costs $c_{ij}$ are estimated within local segments, the final tracking trajectories are obtained through network flow optimization on the whole sequence. After tracklet association, there may exist some gaps between adjacent tracklets in each trajectory due to missed detections and occlusions. The possible gaps in the tracking trajectories are interpolated linearly.

However, in the proposed tracklet affinity model as depicted in Equ. \eqref{eq:17d}, the motion-based affinity model, $P_m(F_i,F_j)$, and appearance-based affinity model, $P_a(F_i,F_j)$, are treated equally without any voting weights. This may result in inaccurate affinity scores if one of the tracking cues is dominant and the other one is confusing, having an effect as a noise factor. Therefore, we investigate the influences of the two tracking cues on tracking performance in next section.

\section{Learning of Affinity Weights}

In difficult situations, where severe occlusions and interactions occur, the motion-based affinity model and appearance-based affinity model may not be consistent. Hence, we need to weight them properly for stable performance.

We propose to add a weighting parameter $\lambda$, which controls the weight of the motion-based affinity score, in Equ. \eqref{eq:17d}.
\vspace*{-0.5\baselineskip}
\begin{align}
\mathcal{S}_{ij}=[{P_m(F_i,F_j)}]^\lambda P_a(F_i,F_j)\mathcal{C}_{ij}, \quad 0 \leq \lambda \leq 1 \label{eq:21a}
\end{align}
where $\lambda$ is learned from labeled data. If the value of $\lambda$ is larger, the motion-based affinity score, ${P_m(F_i,F_j)}$, contributes more to $\mathcal{S}_{ij}$.

\subsection{Assessment of Difficult Situations for Motion Dynamics}

To investigate the difficult situations where the motion affinities are not reliable, we only utilize the motion-based affinity model to estimate $c_{ij}$ for Equ. \eqref{eq:7} in the experiments. After analyzing the inconsistencies between the tracking results and the labeled ground truth data, we obtain these typical situations where the motion affinities are not reliable. Based on the analysis, we design a rule to automatically assess these difficult situations.

There are two constraints in this rule. The first one is that two tracklets should have a certain amount of overlap in the starting or ending frames, which indicates the occlusion of two targets:
\vspace*{-0.5\baselineskip}
\begin{align}
&S(z_i^{t^e})\cap S(z_k^{t^e})\geq \eta*\min(S(z_i^{t^e}),S(z_k^{t^e})) \ \text{or} \notag \\
&S(z_i^{t^s})\cap S(z_k^{t^s})\geq \eta*\min(S(z_i^{t^s}),S(z_k^{t^s})) \label{eq:21b}
\end{align}
where $z_i^{t^e}$ and $z_i^{t^s}$ are the detection responses of $F_i$ at the ending frame $t^e$ and the starting frame $t^s$, respectively. $z_k^{t^e}$ and $z_k^{t^s}$ are the detection responses of $F_k$ at the ending frame $t^e$ and the starting frame $t^s$, respectively. $S(\cdot)$ is the operator to capture the area of the detections in pixels. $\eta$ is a sensitivity threshold ($\eta=0.3$ in our implementation).

The second constraint is that tracklets $F_i$ and $F_j$ must have a gap for it to be linkable. That is,
\vspace*{-0.5\baselineskip}
\begin{align}
t_j^s-t_i^e>1, \quad i\neq j \label{eq:21b2}
\end{align}
where $t_j^{s}$ is the starting frame of tracklet $F_j$, $t_i^{e}$ is the ending frame of tracklet $F_i$, and $t_j^s>t_i^e$.

\begin{figure}[!ht]
\centering
\includegraphics[width=0.635\linewidth]{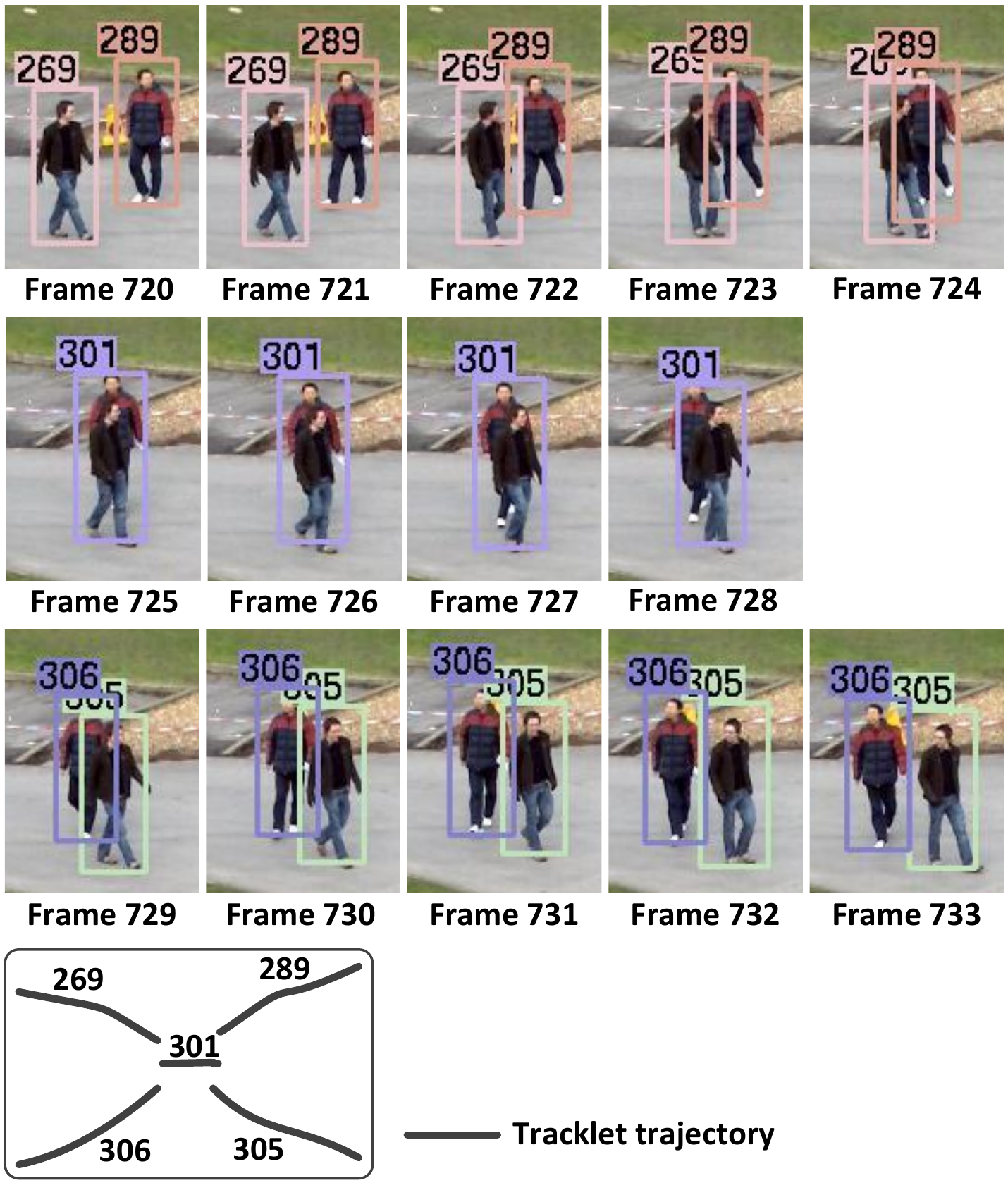}
   \caption{An example of the difficult situations.}
\label{fig:04}
\end{figure}

If the tracklet pair $\{F_i, F_k\}$ match the above two conditions, then we add the weighting parameter $\lambda$ to the motion-based affinity models related to $F_i$ and $F_k$ as $[{P_m(F_i,F_j)}]^\lambda$ and $[{P_m(F_k,F_l)}]^\lambda$. $F_j$, $F_l$ are the linked candidate tracklets of $F_i$ and $F_k$ respectively. As we can see in Figure \ref{fig:04}, the tracklet pairs with identities 269, 289 and 305, 306 match these two constraints. In such situation, due to interactions of the two targets, the two corresponding trajectories become ambiguous. Hence, we add the weighting parameter $\lambda$ for these 4 tracklets in terms of the motion-based affinity models.

\subsection{Learning of the Weighting Parameter}

\begin{algorithm}[t]\footnotesize
\caption{Weighting parameter learning for tracklet association}
\label{alg:01}
\begin{algorithmic}[1]
\REQUIRE ~~ \\
Reliable tracklets;\\
Labeled ground truth data;\\
\ENSURE ~~ \\
The learned weighting parameters: $\{\lambda_1,\lambda_2\}$;
\STATE Initialize the weighting parameters: $\lambda_1={\lambda^l}_1=0,\lambda_2={\lambda^l}_2=0$ and the step value $\Delta \lambda=0.1$;
\STATE Online estimate transition costs for all graph node (tracklet) pairs based on: \\
$c_{ij}=-\log ({P_m(F_i,F_j)} P_a(F_i,F_j)\mathcal{C}_{ij})$;
\FOR {i=1 to 2}
\WHILE {${\lambda^l}_i\leq 1$}
\FORALL {graph node (tracklet) pair $\{F_i, F_k\}$ matches the rule of the assessment of difficult situations}
    \IF {$u\geq 1$ and $u\leq B_1$}
    \STATE $\lambda=\lambda_1$;
    \ELSIF {$u> B_1$}
    \STATE $\lambda=\lambda_2$;
    \ENDIF
\FORALL {tracklet pairs related to $F_i$, $F_k$}
\STATE $c_{ij}=-\log ([{P_m(F_i,F_j)}]^\lambda P_a(F_i,F_j)\mathcal{C}_{ij})$; \\
$c_{kl}=-\log ([{P_m(F_k,F_l)}]^\lambda P_a(F_k,F_l)\mathcal{C}_{kl})$;
\ENDFOR
\ENDFOR
\STATE Obtain tracking results through network flow optimization;
\IF {Current tracking results better converge to ground truth data}
\STATE $\lambda_i={\lambda^l}_i$;
\ENDIF
\STATE ${\lambda^l}_i={\lambda^l}_i+\Delta \lambda$;
\ENDWHILE
\ENDFOR
\RETURN $\{\lambda_1,\lambda_2\}$.
\end{algorithmic}
\end{algorithm}

The weighting parameter $\lambda$ in Equ. \eqref{eq:21a} defines the weight of the motion-based affinity model for tracklet association. Based on the number of frames of the gaps between tracklets, we divide $\lambda$ into 2 levels:
\vspace*{-0.5\baselineskip}
\begin{align}
\lambda = \left\{
 \begin{array}{lll}
             \lambda_1, & 1\leq u \leq B_1  \\
             \lambda_2, & u >B_1
             \end{array}
\right. \label{eq:21c}
\end{align}
where $u$ is the number of frames in the gap between corresponding tracklets ($B_1=20$ in our implementation).

The intuition is that if the gaps are longer, it is more difficult to accurately estimate the joint tracklet dynamics. Therefore, we define 2 difficulty levels as in Equ. \eqref{eq:21c}, making the weighting parameter $\lambda$ adaptive to the difficult situations. The number of difficulty levels and the upper bound value of level 1 ($B_1$) are empirically determined. Furthermore, we employ tracking performance evaluation and network flow optimization jointly to optimize the weighting parameters for tracklet association.

The learning algorithm is summarized in Algorithm \ref{alg:01}. Given the reliable tracklets and the labeled ground truth data of a video sequence, we aim to learn the weighting parameters in a supervised manner so that the tracking performance can be optimized. The two proposed weighting parameters are optimized independently in a greedy fashion. The weighting parameters are initialized as $\lambda_1=0,\lambda_2=0$. After some iterations with a fixed step value, the weighting parameters $\{\lambda_1,\lambda_2\}$ are obtained.

The weighting parameters $\lambda_1$ and $\lambda_2$ are learned from the ground truth data of PETS 2009 \cite{Ferryman} in this paper. Our tracking algorithm is then run with the learned weighting parameters ($\lambda_1=0.5, \lambda_2=0.2$) on all the datasets for evaluation. Based on the analysis of the experimental results with different $B_1$, we find that the tracking performance is slightly affected by the changes of $B_1$. An upper bound value B1 of (20-30) frame gap works well for all sequences.

\section{Experiments}

\subsection{Datasets}
To evaluate the performance of the proposed approach, we experiment on five challenging, publicly available pedestrian datasets.

\textbf{TUD.} The TUD Crossing sequence \cite{Andriluka2} and TUD-Stadtmitte sequence \cite{Andriluka} are real-world videos filmed in busy pedestrian streets. The cameras are positioned at a quite low angle, resulting in more complex occlusion patterns and rather inaccurate ground plane locations. Furthermore, for TUD-Stadtmitte, the size of the pedestrians on the image plane vary drastically.

\textbf{PETS 2009.} This benchmark dataset \cite{Ferryman} presents an outdoor scene with large number of pedestrians captured from multiple cameras at 7 fps. The pedestrians vary significantly in appearance due to shadows and lighting changes. Moreover, there are frequent occlusions, caused by pedestrian occluding each other, or static occlusions such as the traffic sign. In the experiments, we use the sequences S2L1 and S2L2 in the first view, which are widely used in literature.

\textbf{Town Centre.} The Town Centre dataset \cite{Benfold} is captured by a single elevated camera in a busy street. There are 16 pedestrians visible at any time on average, leading to frequent dynamic occlusions and interactions. Furthermore, due to the severe occlusions caused by static obstacles, many pedestrians are not detected by the state-of-the-art detectors.

\textbf{ETH.} The ETH BAHNHOF and SUNNY DAY sequences \cite{Ess} show busy street scenes from a pair of cameras on a moving stroller. The stroller is moving forward at most of the time, however there are still some panning motions, which leads to the unreliable motion affinities between tracklets. Moreover, frequent full or partial occlusions occur due to the low view angles of cameras. The size of the pedestrians also varies significantly on the image plane.

\textbf{MOTChallenge.} The MOTChallenge 2D Benchmark is an up-to-date multiple object tracking benchmark. It consists of a total of 22 sequences, in which half of them are used for training and half of them are used for testing. The test sequences cover many different situations, such as different viewpoints, static or moving camera, different weather conditions. This makes MOTChallenge benchmark very challenging.

\begin{figure*}[!ht]
\centering
\subfigure{
\includegraphics[width=0.195\linewidth]{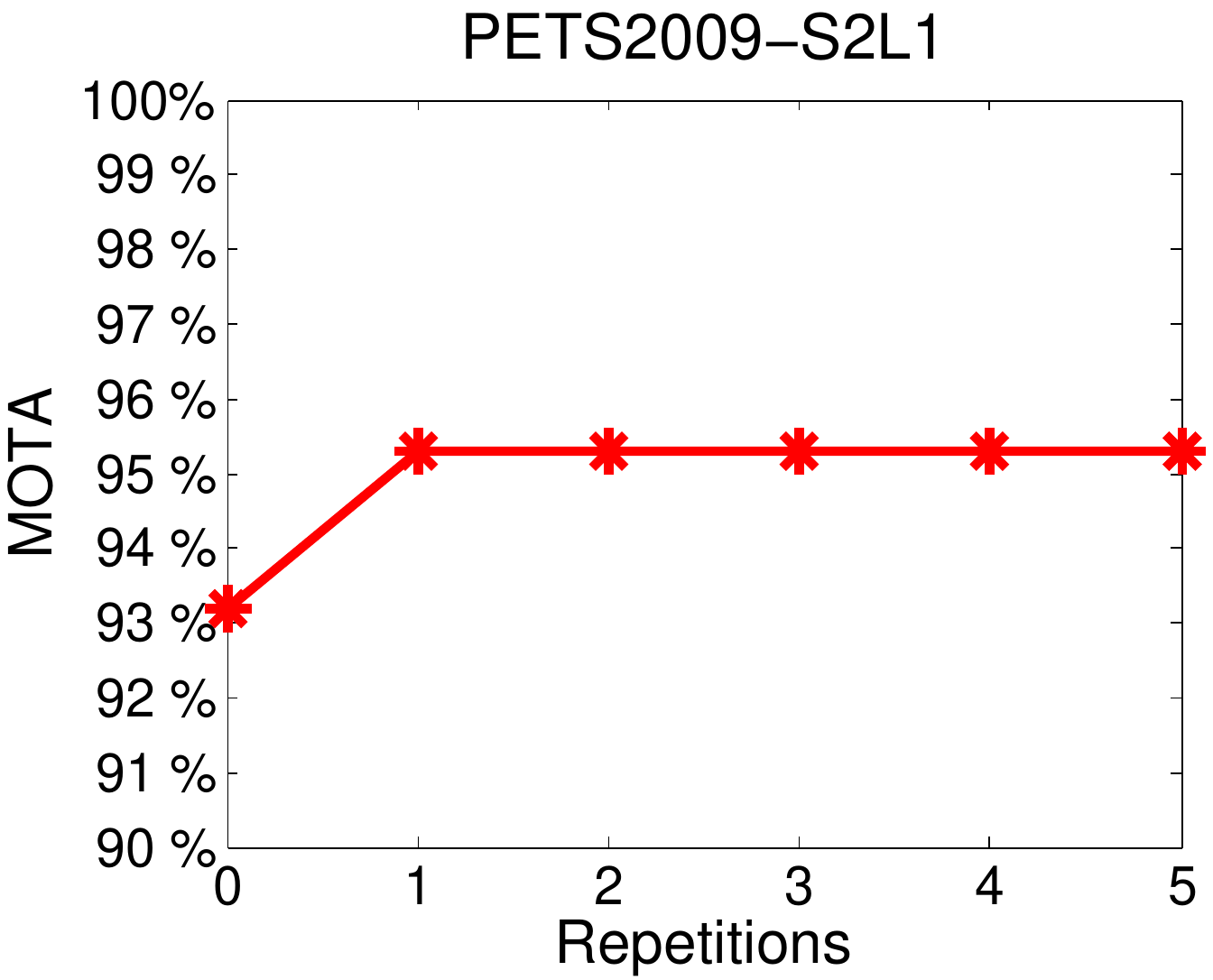}}
\subfigure{
\includegraphics[width=0.195\linewidth]{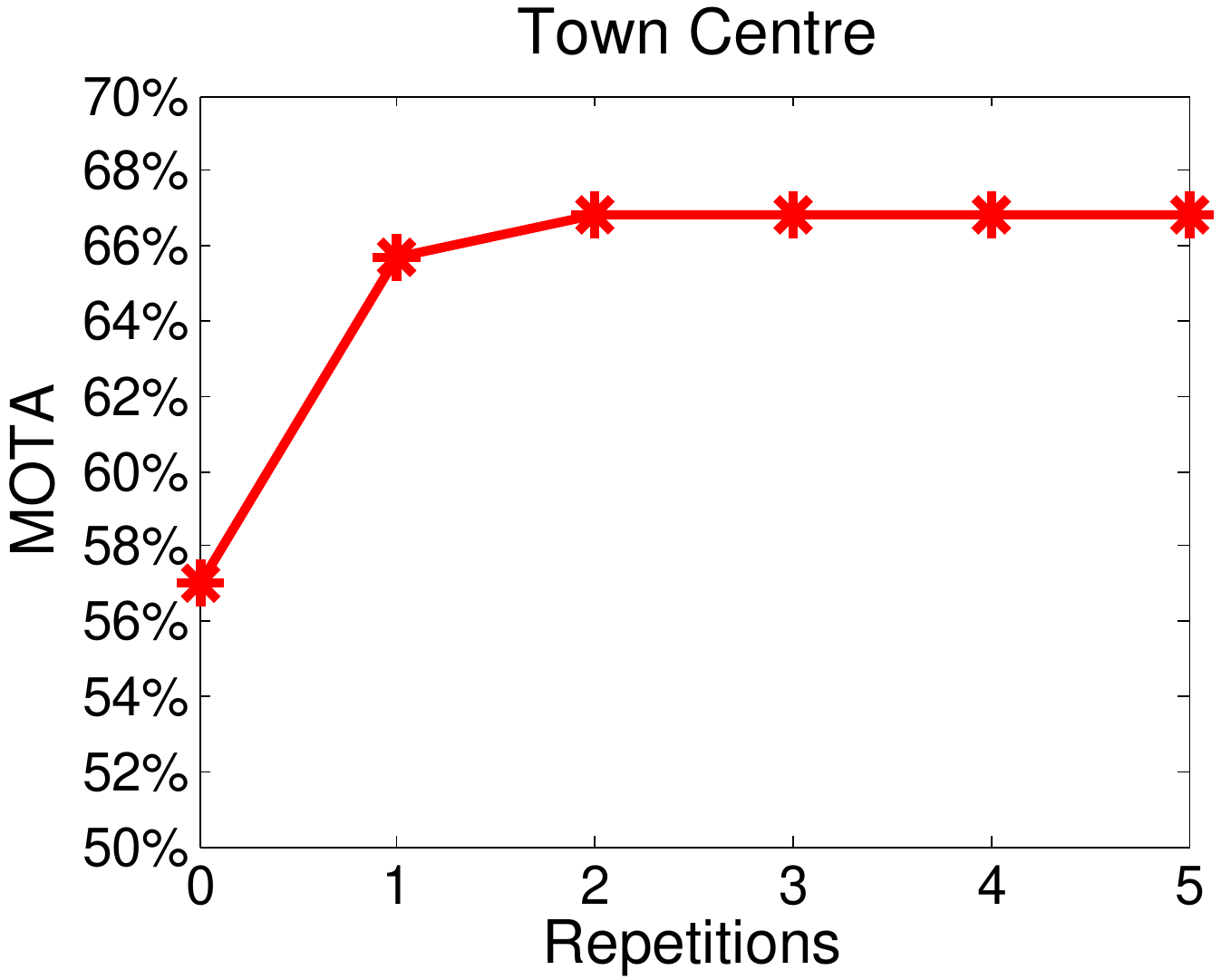}}
\subfigure{
\includegraphics[width=0.195\linewidth]{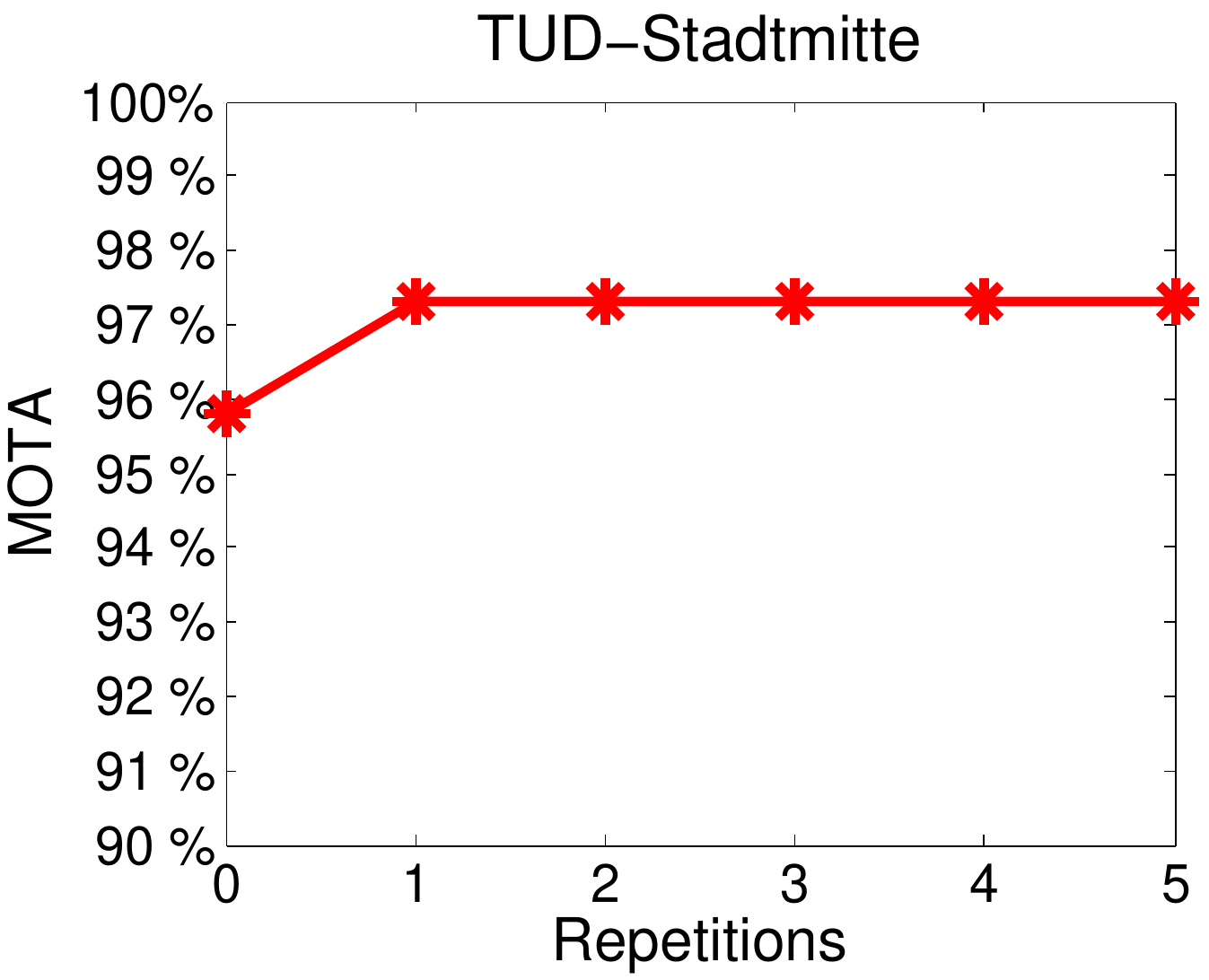}}
\subfigure{
\includegraphics[width=0.195\linewidth]{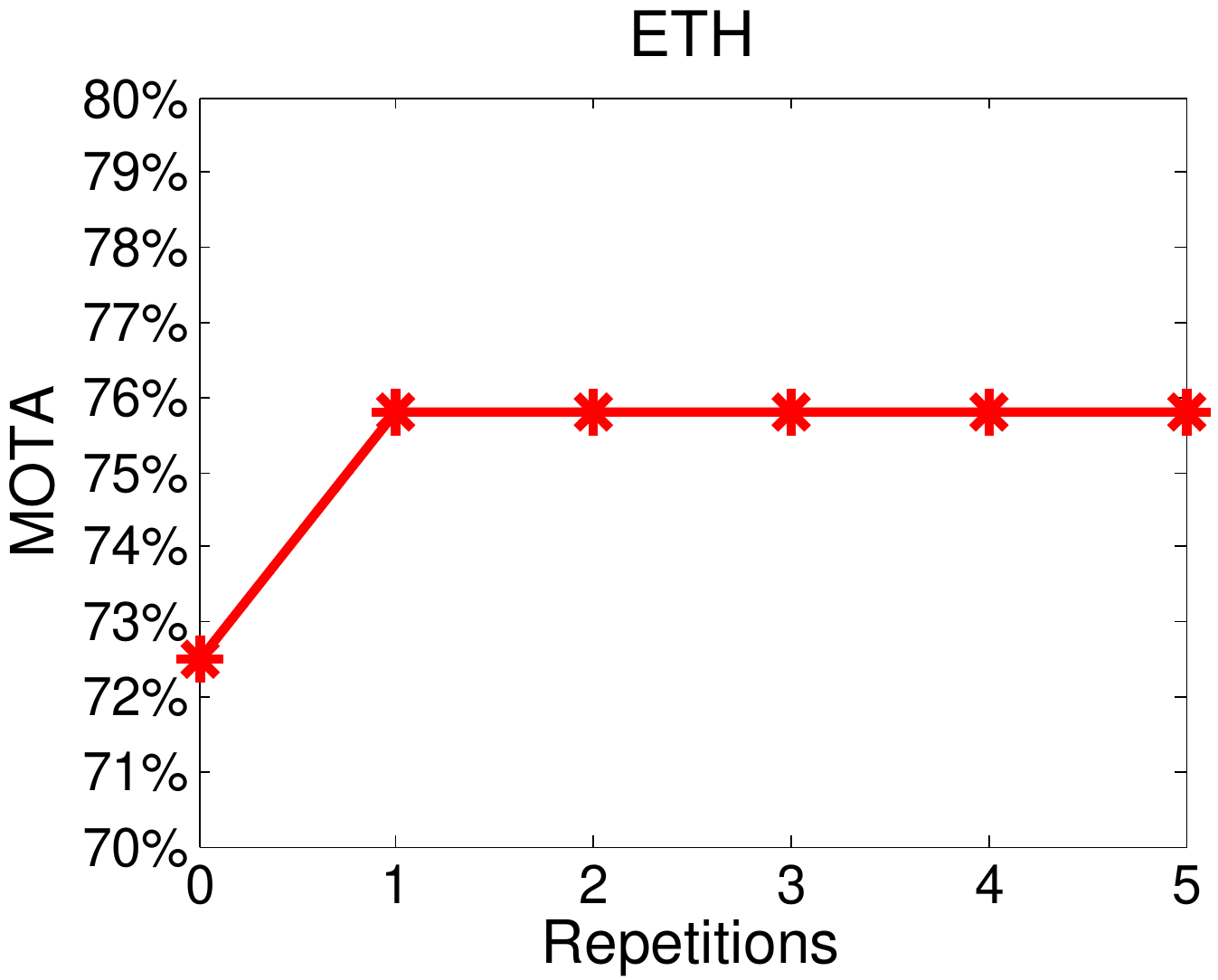}}
\caption{Influence of tracklet refinement process on tracking performance. Each plot presents the performance (measured by MOTA score) on a particular dataset w.r.t. the number of repetitions of tracklet refinement process.}
\label{fig:subfig1} 
\end{figure*}

\subsection{Experimental Settings}

The online collected training samples from video frames are normalized to $128\times 64$ pixels for target-specific metric learning. For the color feature, RGB, YCbCr and HSV color histograms are extracted with 16 bins for each channel respectively and concatenated into a 144-element vector. To capture shape information, we adopt the Histogram of Gradients (HOG) feature \cite{Dalal} by setting the cell size to be 8 to form a 3968-element vector.  Two types of texture features are extracted by Schmid and Gabor filters. In total, 13 Schmid channel features and 8 Gabor channel features are obtained to form a 336-element vector by using a 16-bin histogram vector to represent each channel. Each person image is thus represented by a feature vector in a 4448-dimensional feature space.

\subsection{Evaluation Metrics}

Since it is difficult to use one single score to evaluate multi-target tracking performance, we utilize the evaluation metrics defined in \cite{Li,Butt}, as well as the standard CLEAR MOT metrics \cite{Bernardin}:
\begin{itemize}
\item \emph{MOTA}($\uparrow$):	Multi-object tracking accuracy.
\item \emph{MOTP}($\uparrow$): Multi-object tracking precision.
\item \emph{Recall}($\uparrow$): correctly matched detections / total detections in ground truth.
\item \emph{Precision}($\uparrow$): correctly matched detections / total detections in the tracking result.
\item \emph{FAF}($\downarrow$): number of false alarms per frame.
\item \emph{GT}: number of trajectories in ground truth.
\item \emph{MT}($\uparrow$): number of mostly tracked trajectories.
\item \emph{PT}: number of partially tracked trajectories.
\item \emph{ML}($\downarrow$): number of mostly lost trajectories.
\item \emph{Frag}($\downarrow$): number of fragmentations.
\item \emph{IDS}($\downarrow$): number of id switches.
\item \emph{IDS/correctly matched detections} ($\downarrow$).
\end{itemize}

For evaluation measures with ($\uparrow$), higher scores denote better performance; for evaluation measures with ($\downarrow$), lower scores denote better performance. A tracking bounding box in the result having more than 50\% overlap with the corresponding groundtruth bounding box is considered as true positive. The evaluation codes are downloaded from \cite{web}.

\subsection{Influence of Tracklet Refinement}

To investigate the influence of the repetitions of tracklet refinement process, we run our tracking algorithm and experimented with different number of repetitions while keeping all the other conditions fixed. As shown in Figure \ref{fig:subfig1}, for each dataset, the changes in performance (measured by MOTA score) is plotted against the repetitions of tracklet refinement process. $0$ repetitions in Figure \ref{fig:subfig1} means that no tracklet refinement is utilized for our tracking algorithm. Note that the tracking performance has been improved by exploiting the tracklet refinement for all datasets. Moreover, it is found that the optimal tracking performance is achieved by no more than $2$ repetitions of tracklet refinement process for the four exemplar datasets. From the statistics as shown in Figure \ref{fig:subfig1}, we can conclude that the proposed tracking algorithm with only one time tracklet refinement process can achieve near-optimal tracking performance.

\subsection{Quantitative Evaluation}

The quantitative evaluations are presented in three sub-sections: abbreviations of the proposed methods in the experiments, comparison with network flow based methods, and comparison with other state-of-the-art methods on benchmark datasets. The tracking results of other methods are extracted from the published papers and the
MOTChallenge benchmark website \cite{MOTChallenge} for the ease of reference.

\subsubsection{Abbreviations of Different Methods}
\begin{itemize}
\item \textbf{\emph{CML:}} The proposed method with only an online learned common class metric for all tracklets.
\item \textbf{\emph{TSML:}} The proposed method with online target-specific metric learning.
\item \textbf{\emph{TD:}} The proposed method with only tracklet dynamics.
\item \textbf{\emph{TSML+TD:}} The proposed method with online target-specific metric learning and tracklet dynamics.
\item \textbf{\emph{TSML+TD+WP:}} The proposed method with full tracklet affinity model including target-specific metric learning (\textbf{TSML}), tracklet dynamics (\textbf{TD}) and weighting parameters (\textbf{WP}).
\end{itemize}

\textbf{CML} and \textbf{TSML} are from our previous work \cite{Wang111}.

\subsubsection{Comparison with Network Flow based Methods}

We first evaluate our method on the popular TUD Crossing sequence \cite{Andriluka2} and ETH BAHNHOF sequence \cite{Ess}. For a fair comparison, we use the same sequences and pre-trained pedestrian detector as used in \cite{Butt}. The quantitative metric that we use is \emph{ID switches / total number of correct observations used in the trajectories (IDS / correctly matched detections)}, which is the same as in \cite{Butt}. Table \ref{tb:1} gives the quantitative results computed on the TUD Crossing sequence, and the first 350 frames of the ETH BAHNHOF sequence. Due to the forward and panning motions of the cameras, the motion affinities between tracklets are unreliable for the BAHNHOF sequence. Therefore, we do not add the online tracklet dynamics estimation for the experiments in this subsection. We report the results of five methods: DP algorithm of \cite{Pirsiavash} is our baseline work without adding the online learning of tracklet affinity models. The results are also compared to those of other network flow based methods: MCNF \cite{Zhang} and LRMCNF \cite{Butt}. The proposed method with online target-specific metric learning is denoted as TSML. Furthermore, we also report the results of the proposed method with only a common class metric for all tracklets, which is denoted as CML.

\begin{table}[!ht]
\centering
\footnotesize
\resizebox{\columnwidth}{!}{%
\begin{tabular}{|c|c|c||c|}
\hline
\ Algorithm & TUD Crossing & ETH & ETH (GT) \\
\hline
$\textbf{DP}$ \cite{Pirsiavash} & 32/768 & 37/1387 & 25/1648 \\
\hline
MCNF \cite{Zhang} & 9/433 & 11/1057 & 5/922 \\
\hline
LRMCNF \cite{Butt} & 14/819 & 23/1514 & 14/1783 \\
\hline
CML & 10/845 & 5/1728 & 3/1786 \\
\hline
TSML & 7/862 & 1/1790 & 0/1820 \\
\hline
\end{tabular}}
\caption{Comparison of tracking results with network flow based methods on TUD Crossing and ETH BAHNHOF (first 350 frames) sequences. The entries in the table are (IDS)/(correctly matched detections). Columns 1 and 2 use the pre-trained human detector of \cite{Felzenszwalb2}. Column 3 shows the results when ground truth bounding boxes are used to generate the initial tracklets. The ground truth bounding boxes are from \cite{ETHdataset}.}
\label{tb:1}
\end{table}

Note that our method gives better results when compared with the three network flow methods \cite{Pirsiavash,Zhang,Butt}. Moreover, the noticeable improvement in ID switches indicates that our method can better deal with long-term tracking, where the traditional motion models are less reliable.

\subsubsection{Comparison with State-of-the-art Methods}

To show the effectiveness of our method, we further compare our method with other state-of-the-art methods on more publicly available datasets. We use the pre-trained human detector of \cite{Felzenszwalb2} to generate the detections.
For the MOTChallenge benchmark \cite{Leal-Taixe03}, we utilize DPM detections \cite{Felzenszwalb2} and the public detections from this benchmark for the evaluation.
In the result tables, the numbers ranked in the first place of the respective evaluation measures are marked in bold.

\begin{table*}[!ht]\footnotesize
\centering
\begin{tabular}{|l|c|c|c|c|c|c|c|c|c|c|c|}
\hline
Method & MOTA & MOTP & Recall & Precision & FAF & GT & MT & PT & ML & Frag & IDS \\
\hline
Energy Minimization \cite{Andriyenko2} & 81.4\% & 76.1\% & - & - & - & 19 & 82.6\% & 17.4\% & \textbf{0.0\%} & 21 & 15 \\
DC Tracking \cite{Andriyenko} & 95.9\% & 78.7\% & - & - & - & 19 & \textbf{100.0\%} & 0.0\% & \textbf{0.0\%} & 8 & 10 \\
KSP \cite{Berclaz} & 80.3\% & 72.0\% & - & - & - & 19 & 73.9\% & 17.4\% & 8.7\% & 22 & 13 \\
MTMM \cite{Henriques01} & 83.3\% & 71.1\% & - & - & - & 19 & 89.5\% & 10.5\% & \textbf{0.0\%} & 45 & 19 \\
UHMTGDA \cite{Hofmann01} & 97.8\% & 75.3\% & - & - & - & 19 & \textbf{100.0\%} & 0.0\% & \textbf{0.0\%} & 8 & 8 \\
HJMRMT \cite{Hofmann02} & 98.0\% & 82.8\% & - & - & - & 19 & \textbf{100.0\%} & 0.0\% & \textbf{0.0\%} & 11 & 10 \\
\textsc{$(MP)^{2}T$} \cite{Izadinia} & 90.7\% & 76.0\% & - & - & - & 19 & - & - & - & - & - \\
DTLE Tracking \cite{Milan} & 90.3\% & 74.3\% & - & - & - & 19 & 78.3\% & 21.7\% & \textbf{0.0\%} & 15 & 22 \\
CEMMT \cite{Milan02} & 90.6\% & 80.2\% & - & - & - & 19 & 91.3\% & 4.4\% & 4.3\% & \textbf{6} & 11 \\
GMCP-Tracker \cite{Zamir01} & 90.3\% & 69.0\% & - & - & - & 19 & 89.5\% & 10.5\% & \textbf{0.0\%} & 54 & 10 \\
OMTD \cite{Breitenstein} & 79.7\% & 56.3\% & - & - & - & - & - & - & - & - & - \\
OMAT \cite{Wu01} & 92.8\% & 74.3\% & - & - & - & 19 & \textbf{100.0\%} & 0.0\% & \textbf{0.0\%} & 11 & 8 \\
PMPTCS \cite{Yang001} & 76.0\% & 53.8\% & - & - & - & - & - & - & - & - & - \\
OGOMT \cite{Possegger} & \textbf{98.1\%} & 80.5\% & - & - & - & 19 & \textbf{100.0\%} & 0.0\% & \textbf{0.0\%} & 16 & 9 \\
CSL-VOX \cite{Chen} & 89.78\% & - & \textbf{98.28\%} & 91.07\% & - & 19 & - & - & - & - & 6 \\
CSL-DPT \cite{Chen} & 88.13\% & - & 97.64\% & 90.45\% & - & 19 & - & - & - & - & 8 \\
\hline
CML & 92.1\% & \textbf{86.4\%} &  95.1\% & 97.6\% & 0.14 & 19 & 94.7\% & 5.3\% & \textbf{0.0\%} & 26 & 28 \\
TSML & 93.4\% & \textbf{86.4\%} & 96.0\% & 97.7\% & 0.13 & 19 & 94.7\% & 5.3\% & \textbf{0.0\%} & 21 & 18 \\
TD & 93.7\% & 86.3\% & 96.6\% & 97.4\% & 0.15 & 19 & 94.7\% & 5.3\% & \textbf{0.0\%} & 17 & 13 \\
TSML+TD & 94.7\% & \textbf{86.4\%} & 97.2\% & 97.6\% & 0.14 & 19 & 94.7\% & 5.3\% & \textbf{0.0\%} & 12 & 7 \\
TSML+TD+WP & 95.3\% & \textbf{86.4\%} & 97.4\% & \textbf{98.0\%} & \textbf{0.11} & 19 & 94.7\% & 5.3\% & \textbf{0.0\%} & 11 & \textbf{4} \\
\hline
\end{tabular} 
\caption{Comparison of tracking results between state-of-the-art methods and ours on PETS2009-S2L1.}
\label{tb:2}
\end{table*}

\begin{table*}[!ht]\footnotesize
\centering
\begin{tabular}{|l|c|c|c|c|c|c|c|c|c|c|c|}
\hline
Method & MOTA & MOTP & Recall & Precision & FAF & GT & MT & PT & ML & Frag & IDS \\
\hline
\textsc{$(MP)^{2}T$} \cite{Izadinia} & \textbf{75.7\%} & 71.6\% & - & - & - & 231 & - & - & - & - & - \\
SGB Tracker \cite{Leal-Taixe} & 71.3\% & 71.8\% & - & - & - & 231 & 58.6\% & 34.4\% & 7.0\% & 363 & 165 \\
GMCP-Tracker \cite{Zamir01} & 75.6\% & 71.9\% & - & - & - & 231 & - & - & - & - & - \\
MCNF \cite{Zhang} & 69.1\% & 72.0\% & - & - & - & 231 & 53.0\% & 37.7 & 9.3\% & 440 & 243 \\
SMT \cite{Benfold} & 64.3\% & \textbf{80.2\%} & - & - & - & 231 & \textbf{67.4\%} & 26.1\% & \textbf{6.5\%} & 343 & 222 \\
MSBMT \cite{Pellegrini} & 65.5\% & 71.8\% & - & - & - & 231 & 59.1\% & 33.9\% & 7.0\% & 499 & 288 \\
OMAT \cite{Wu01} & 69.5\% & 68.7\% & - & - & - & 231 & 64.7\% & 27.4\% & 7.9\% & 453 & 209 \\
WAYWAG \cite{Yamaguchi} & 66.6\% & 71.7\% & - & - & - & 231 & 58.1\% & 35.4\% & \textbf{6.5\%} & 492 & 302 \\
OGOMT \cite{Possegger} & 70.7\% & 68.6\% & - & - & - & 231 & 56.3\% & 36.3\% & 7.4\% & 321 & \textbf{157} \\
\hline
CML & 55.3\% & 72.6\% &  69.9\% &  86.5\% & 1.79 & 231 & 52.8\% & 36.4\% & 10.8\% & 508 & 327\\
TSML & 57.3\% & 72.9\% & 72.3\% & 87.5\% & 1.58 & 231 & 60.2\% & 30.7\% & 9.1\% & 326 & 269 \\
TD & 55.7\% & 73.2\% & 71.5\% & 86.9\% & 1.71 & 231 & 55.8\% & 34.2\% & 10.0\% & 362 & 264 \\
TSML+TD & 61.6\% & 74.3\% & 74.0\% & 89.5\% & 1.38 & 231 & 61.9\% & 29.9\% & 8.2\% & 259 & 214 \\
TSML+TD+WP & 66.8\% & 74.4\% & \textbf{75.2\%} & \textbf{92.5\%} & \textbf{0.96} & 231 & 64.9\% & 28.2\% & 6.9\% & \textbf{198} & 162 \\
\hline
\end{tabular}
\caption{Comparison of tracking results between state-of-the-art methods and ours on Town Centre dataset.}
\label{tb:3}
\end{table*}

\textbf{PETS2009-S2L1.} For a fair comparison, we utilize the same ground truth as in \cite{Milan02} for the experiments, in which all the occurring pedestrians have been annotated. The quantitative results are shown in Table \ref{tb:2}. As expected, taking tracklet dynamics into account increases the overall tracking performance. Our full tracklet affinity model (TSML+TD+WP) further raises the MOTA by $0.6\%$ and reduces the ID switches by $\approx 43\%$. This indicates that our method with full tracklet affinity model combines motion and appearance cues properly, resulting in further improvement on tracking performance.
On the whole, our method with full tracklet affinity model achieves the best performance compared with 16 state-of-the-art methods in terms of MOTP, Precision, FAF, ML and IDS. For other evaluation measures, our approach also achieves comparative performance.

\textbf{Town Centre.} To show the generality of the learned weighting parameters, we evaluate our approach with the learned weighting parameters on Town Centre dataset. The ground truth we used here is provided by \cite{Benfold}, which is the same as in the compared methods. Due to severe occlusions caused by static obstacles (such as benches) and more frequent dynamic interactions between pedestrians, many pedestrians cannot be detected by the state-of-the-art detectors. Hence, the Recall (as shown in Table \ref{tb:3}) is lower than the other datasets. As we can see in Table \ref{tb:3}, the full tracklet affinity model with the weighting parameters achieves better or nearly the same performances on all evaluation items. Compared with the tracklet affinity model without weighting parameters, the MOTA is improved by about $8.4\%$; recall and precision are improved by about $1.6\%$ and $3.4\%$ respectively; fragments and ID switches are reduced by $23.6\%$ and $24.3\%$ respectively. The obvious improvements in performance indicate that the learned weighting parameters are applicable to new data.

\textbf{TUD-Stadtmitte.} To make a fair comparison, the experiments are conducted using the same ground truth as defined in \cite{Yang2}. The quantitative tracking results are shown in Table \ref{tb:4}. Though there are occlusions and interactions between pedestrians, the number of pedestrians appearing in the scene is less than other datasets. Hence, our method can generate better optimal tracking results on TUD-Stadtmitte than other datasets. Note that our proposed method with target-specific metric learning (TSML) has achieved very good performance. We also provide the tracking results of the proposed method with TSML+TD and TSML+TD+WP, which show the same optimal tracking results as shown in Table \ref{tb:4}. Compared with \cite{Kuo,Yang2}, the improvement is obvious for some metrics. Our method achieves the highest recall and the mostly tracked score (MT) among all the methods. It also achieves the lowest ID switches. Meanwhile, our method achieves competitive performance on precision, false alarms per frame and fragments compared with \cite{Kuo,Yang2}.

\begin{table*}[!ht]\footnotesize
\centering
\begin{tabular}{|l|c|c|c|c|c|c|c|c|c|c|c|}
\hline
Method & MOTA & MOTP & Recall & Precision & FAF & GT & MT & PT & ML & Frag & IDS \\
\hline
Energy Minimization \cite{Andriyenko2} & - & - & - & - & - & 9 & 60.0\% & 30.0\% & \textbf{0.0\%} & 4 & 7 \\
DC Tracking \cite{Andriyenko} & - & - & 74.7\% & 84.2\% & 0.870 & 10 & 50.0\% & 50.0\% & \textbf{0.0\%} & 8 & 10 \\
PRIMPT \cite{Kuo} & - & - & 81.0\% & \textbf{99.5\%} & \textbf{0.028} & 10 & 60.0\% & 30.0\% & 10.0\% & \textbf{0} & 1 \\
Online CRF Tracking \cite{Yang2} & - & - & 87.0\% & 96.7\% & 0.184 & 10 & 70.0\% & 30.0\% & \textbf{0.0\%} & 1 & \textbf{0} \\
\hline
CML & 94.5\% & \textbf{72.7\%} & 95.1\% & 99.4\% & 0.030 & 10 & \textbf{100\%} & 0.0\% & \textbf{0.0\%} & 2 & 1 \\
TSML & \textbf{97.3\%} & 71.3\% & \textbf{98.0\%} & 99.3\% & 0.040 & 10 & \textbf{100\%} & 0.0\% & \textbf{0.0\%} & 3 & \textbf{0} \\
TSML+TD & \textbf{97.3\%} & 71.3\% & \textbf{98.0\%} & 99.3\% & 0.040 & 10 & \textbf{100\%} & 0.0\% & \textbf{0.0\%} & 3 & \textbf{0} \\
TSML+TD+WP & \textbf{97.3\%} & 71.3\% & \textbf{98.0\%} & 99.3\% & 0.040 & 10 & \textbf{100\%} & 0.0\% & \textbf{0.0\%} & 3 & \textbf{0} \\
\hline
\end{tabular}
\caption{Comparison of tracking results between state-of-the-art methods and ours on TUD Stadtmitte dataset.}
\label{tb:4}
\end{table*}

\textbf{ETH.} To see the effectiveness of the proposed method, we further evaluate it on the challenging ETH dataset \cite{Ess}. Due to the unreliable motion affinities between tracklets of this dataset, we use the tracklet affinity model without TD and WP for the experiments. For a fair comparison, we use the ground truth provided by \cite{Yang2}. The quantitative tracking results are shown in Table \ref{tb:5}. We can see that our method can achieve better or competitive performance on all the commonly used evaluation measures. Compared with \cite{Kuo}, the most related work, the recall and precision are improved by 4.1\% and 7.6\% respectively; the MT is improved by 7.2\%; false alarms per frame are reduced by 41.6\%; and ID switches are reduced by 54.5\%. The significant reduction in ID switches and false alarms indicates that our target-specific appearance-based model is superior to the method by \cite{Kuo}.

\begin{table*}[!ht]\footnotesize
\centering
\begin{tabular}{|l|c|c|c|c|c|c|c|c|c|c|c|}
\hline
Method & MOTA & MOTP & Recall & Precision & FAF & GT & MT & PT & ML & Frag & IDS \\
\hline
PRIMPT \cite{Kuo} & - & - & 76.8\% & 86.6\% & 0.891 & 125 & 58.4\% & 33.6\% & 8.0\% & 23 & 11 \\
Online CRF Tracking \cite{Yang2} & - & - & 79.0\% & 90.4\% & 0.637 & 125 & 68.0\% & 24.8\% & 7.2\% & \textbf{19} & 11 \\
DTLE Tracking \cite{Milan} & - & - & 77.3\% & 87.2\% & - & 125 & 66.4\% & 25.4\% & 8.2\% & 69 & 57 \\
LIMC Tracking \cite{Leal-Taixe02} & - & - & \textbf{83.8\%} & 79.7\% & - & 125 & \textbf{72.0\%} & 23.3\% & \textbf{4.7\%} & 85 & 71 \\
\hline
CML & 71.6\% & 77.0\% & 78.7\% & 92.0\% & 0.710 & 125 & 60.0\% & 29.6\% & 10.4\% & 77 & 19 \\
TSML & \textbf{75.8\%} & \textbf{77.2\%} & 80.9\% & \textbf{94.2\%} & \textbf{0.520} & 125 & 65.6\% & 24.0\% & 10.4\% & 26 & \textbf{5} \\
\hline
\end{tabular}
\caption{Comparison of tracking results between state-of-the-art methods and ours on ETH dataset. Note that the forward and panning motions of the cameras lead to unreliable motion affinities between tracklets of this dataset. We thus do not employ TD and WP in the experiment.}
\label{tb:5}
\end{table*}

\textbf{MOTChallenge 2D Benchmark.} To further show more meaningful quantitative evaluation of the proposed method, we evaluate it on the recent MOTChallenge 2D Benchmark \cite{Leal-Taixe03}. The number of test sequences in this benchmark is 11, in which 5 of them are taken by moving cameras and 6 of them are taken by static cameras. For the sequences taken by static cameras, our full tracklet affinity model (TSML+TD+WP) is used for evaluation. Due to the moving cameras, the motion affinities estimated by proposed tracklet dynamic model are not reliable. Hence, for the sequences taken by moving cameras, our proposed method with only target-specific metric learning (TSML) is used for evaluation.
To further show the generalization capability of the learned weighting parameters, the same parameters $\lambda_1=0.5, \lambda_2=0.2$ are used for all the 6 testing sequences taken by static cameras. The evaluation results are generated from the MOTChallenge benchmark website \cite{MOTChallenge}. Hence, only the optimal results of the proposed method are provided. As shown in Table \ref{tb:6}, compared with other state-of-the-art methods, our method achieves better or competitive performance on all the evaluation measures.

\textbf{PETS2009-S2L2.} To show the effective of the proposed method on more crowded sequences in PETS 2009 dataset, we further evaluate our method on PETS2009-S2L2 sequence. The evaluation result is generated from the MOTChallenge benchmark website \cite{MOTChallenge}. As shown in Table \ref{tb:7}, our method achieves the best performance on MOTA, MOTP, MT and ML compared with other state-of-the-art methods. For other evaluation items, our method also achieves competitive performance.

\begin{table*}[!ht]
\footnotesize
\centering
\begin{tabular}{|l|c|c|c|c|c|c|c|c|c|}
\hline
Method & MOTA & MOTP & FAF & GT & MT & PT & ML & Frag & IDS \\
\hline
DP\_NMS \cite{Pirsiavash} & 14.5\% & 70.8\% & 2.3 & 721 & 6.0\% & 53.2\% & 40.8\% & 3090 & 4537 \\
TC\_ODAL \cite{Bae} & 15.1\% & 70.5\% & 2.2 & 721 & 3.2\% & 41\% & 55.8\% & 1716 & 637 \\
CEM \cite{Milan02} & 19.3\% & 70.7\% & 2.5 & 721 & 8.5\% & 45\% & 46.5\% & 1023 & 813 \\
SMOT \cite{Dicle} & 18.2\% & 71.2\% & 1.5 & 721 & 2.8\% & 42.4\% & 54.8\% & 2132 & 1148 \\
TBD \cite{Geiger}& 15.9\% & 70.9\% & 2.6 & 721 & 6.4\% & 45.7\% & 47.9\% & 1963 & 1939 \\
LP2D \cite{Leal-Taixe03} & 19.8\% & 71.2\% & 2.0 & 721 & 6.7\% & 52.1\% & 41.2\% & 1712 & 1649 \\
RMOT \cite{Yoon} & 18.6\% & 69.6\% & 2.2 & 721 & 5.3\% & 41.4\% & 53.3\% & 1282 & 684 \\
\hline
Ours1 & \textbf{49.1\%} & \textbf{74.3\%} & \textbf{0.9} & 721 & \textbf{30.4\%} & 43.2\% & \textbf{26.4\%} & 1034 & 637 \\
Ours2 & 34.3\% & 71.7\% & 1.4 & 721 & 14.0\% & 46.6\% & 39.4\% & \textbf{959} & \textbf{618} \\
\hline
\end{tabular}
\caption{Comparison of tracking results between state-of-the-art methods and ours on MOTChallenge 2D Benchmark. ``Our1" uses the DPM detections \cite{Felzenszwalb2}. ``Ours2" uses the public detections from this benchmark \cite{MOTChallenge}.}
\label{tb:6}
\end{table*}

\begin{table*}[!ht]
\footnotesize
\centering
\begin{tabular}{|l|c|c|c|c|c|c|c|c|c|}
\hline
Method & MOTA & MOTP & FAF & GT & MT & PT & ML & Frag & IDS \\
\hline
DP\_NMS \cite{Pirsiavash} & 33.8\% & 69.4\% & 2.2 & 42 & 7.1\% & 83.4\% & 9.5\% & 705 & 1029 \\
TC\_ODAL \cite{Bae} & 30.2\% & 69.2\% & 2.5 & 42 & 2.4\% & 78.6\% & 19.0\% & 499 & 284 \\
CEM \cite{Milan02} & 44.9\% & 70.2\% & 1.5 & 42 & 11.9\% & 73.8\% & 14.3\% & \textbf{165} & \textbf{150} \\
SMOT \cite{Dicle} & 34.4\% & 70.0\% & \textbf{1.1} & 42 & 0.0\% & 76.2\% & 23.8\% & 514 & 251 \\
TBD \cite{Geiger} & 35.5\% & 69.2\% & 3.1 & 42 & 7.1\% & 78.6\% & 14.3\% & 480 & 523 \\
LP2D \cite{Leal-Taixe03} & 40.7\% & 70.2\% & 1.9 & 42 & 9.5\% & 73.8\% & 16.7\% & 359 & 319 \\
RMOT \cite{Yoon} & 37.2\% & 67.7\% & 2.6 & 42 & 9.5\% & 76.2\% & 14.3\% & 320 & 190 \\
\hline
Ours1 & \textbf{59.7\%} & \textbf{74.4\%} & 2.3 & 42 & \textbf{31.0\%} & 64.2\% & \textbf{4.8\%} & 200 & 173 \\
Ours2 & 51.5\% & 70.6\% & 2.1 & 42 & 14.3\% & 76.2\% & 9.5\% & 198 & 165 \\	
\hline
\end{tabular}
\caption{Comparison of tracking results between state-of-the-art methods and ours on PETS2009-S2L2. ``Our1" uses the DPM detections \cite{Felzenszwalb2}. ``Ours2" uses the public detections from \cite{MOTChallenge}.}
\label{tb:7}
\end{table*}

\subsection{Computational Speed}

The computation speed depends on the number of targets in a video sequence. Our method is implemented using MATLAB on a 3.3 GHz, 4 core PC with 8 GB memory. The speed of the proposed method with target-specific metric learning (TSML) is about 13, 6, 10 and 9 fps for PETS 2009, Town Centre, TUD and ETH datasets, respectively, excluding the detection time; for PETS 2009 and Town Centre datasets, the speed of the proposed method with our full tracklet affinity model is 11 and 5 fps respectively. The average speed of the proposed method on MOTChallenge 2D Benchmark is about 7 fps. The online learning of tracklet affinity models is the most time consuming part of our method, which takes up about 90\% of the total computation time. The breakdown is as follows: the learning of appearance-based tracklet affinity model and motion dynamics estimation take up about 80\% and 10\% respectively. Speed-up can be achieved by parallel implementations of the online learning of target-specific metrics. Furthermore, the learning of appearance-based tracklet affinity model and motion dynamics estimation can also be implemented in parallel.

\section{Conclusion}

We have presented our method developed for tracking multiple objects of interest in the scene over a longer period with the aim to maintain consistent tracking and tagging of objects, reducing identity switches. Our method processes the initial tracklets (track fragments) produced by a simple trajectory based tracking algorithm. We propose a two-step online target-specific metric learning to improve the similarity measure based on the appearance cues, and together with coherent dynamics estimation for tracklets based on the motion cues, we establish our new affinity model. The tracking of objects is accomplished by performing tracklet association with network flow optimization where the nodes in the network are tracklets. Thus, our proposed method exploiting both appearance and motion cues is capable to prevent identity switches during tracking and recover missed detections.
Our method is found to be effective even when the appearance or motion cues fail to identify or follow the target due to occlusions or object-to-object interactions. To further improve our method, we also propose to learn the weights of these two tracking cues in our affinity model. Our tracking algorithm has been validated on several public datasets and the experimental results show that it outperforms several state-of-the-art tracking algorithms.

\section*{Acknowledgments}

The authors would like to acknowledge the Research Scholarship from School of EEE, Nanyang Technological University, Singapore.

\ifCLASSOPTIONcaptionsoff
  \newpage
\fi


{\small
\footnotesize
\bibliographystyle{ieee}
\bibliography{egbib}
}

\end{document}